\documentclass[10pt,twocolumn,letterpaper]{article}

\usepackage{cvpr}
\usepackage{times}
\usepackage{epsfig}
\usepackage{graphicx}
\usepackage{amsmath}
\usepackage{amssymb}
\usepackage[ruled,commentsnumbered]{algorithm2e}

\cvprfinalcopy 

\def\cvprPaperID{634} 

\begin{document}

\title{Face Verification Using Boosted Cross-Image Features}


\author{Dong Zhang\\
University of Central Florida\\
Orlando, FL\\
{\tt\small dzhang@cs.ucf.edu}
\and
Omar Oreifej\\
University of California, Berkeley\\
Berkeley, CA\\
{\tt\small oreifej@eecs.berkeley.edu}
\and
Mubarak Shah\\
University of Central Florida\\
Orlando, FL\\
{\tt\small shah@crcv.ucf.edu}
}

\maketitle

\begin{abstract}
This paper proposes a new approach for face verification, where a
pair of images needs to be classified as belonging to the same
person or not. This problem is relatively new and not
well-explored in the literature. Current methods mostly adopt
techniques borrowed from face recognition, and process each of the
images in the pair independently, which is counter intuitive. In
contrast, we propose to extract cross-image features, i.e.
features across the pair of images, which, as we demonstrate, is
more discriminative to the similarity and the dissimilarity of
faces. Our features are derived from the popular Haar-like
features, however, extended to handle the face verification
problem instead of face detection. We collect a large bank of
cross-image features using filters of different sizes, locations,
and orientations. Consequently, we use AdaBoost to select and
weight the most discriminative features. We carried out extensive
experiments on the proposed ideas using three standard face
verification datasets, and obtained promising results
outperforming state-of-the-art.
\end{abstract}

%
%

\section{Introduction}
Facial image analysis is a widely investigated area in computer
vision and multimedia, with several face-relevant problems
explored in the literature. In face detection
\cite{Rowley1998,Hsu2002,Tsao2010}, an image is classified as a
face or not by capturing the facial features and landmarks which
distinguishes a face from other types of objects. On the other
hand, in face recognition
\cite{Turk1991,Wiskott1997,Wright2009,Chen2011}, given an image of
a face, possibly detected using a face detector, the person's
identity is recognized by capturing facial features which
distinguish individuals.

Lately, a new face-relevant problem has been receiving increasing
attention, namely, face verification. Given a pair of face images,
the task is to verify whether they belong to the same person or
not. Face verification is still poorly addressed with only few
reported results. In \cite{Guillaumin2010}, Guillaumin et al.
employed a metric learning method to obtain the optimal
Mahalanobis metrics over a given representation space of faces. On
the other hand, in \cite{Schroff2011}, Schroff et al. proposed a
new similarity measure which increased the verification robustness
against pose, illumination, and expression differences. Also in
\cite{Wolf2011}, a SVM-based method was proposed for video face
verification, where a SVM is trained on a video of a person, then
used to classify another video in order to determine if it
corresponds to the same identity. Moreover, Kumar et al.
\cite{Kumar2009} successfully employed attribute classifiers for
face verification which significantly improved the results of face
verification. Also Nguyen et al. \cite{Nguyen2011} used an
adaptive learning method to obtain the optimal transformation for
the cosine similarity metric in the face space. Additionally, the
Wavelet LBP features were proposed in \cite{Goh2011} and employed
for face verification. It can be clearly observed that current
face verification methods mostly employ techniques borrowed from
face recognition and apply them separately on each image of the
pair to be verified.

In this paper, we propose a novel method for face verification,
which we believe will set a new direction in this problem. Our
method is inspired by the following observation: Since face
verification is a two-image problem, using features extracted
individually from each image is counter intuitive; rather, the
most discriminative features should be extracted across the pair
of images jointly. Methods such as \cite{Georghiades2001} and
\cite{Everingham2006} employ LBP and SIFT features which are
designed to describe a certain region in a single image. In
contrast, we propose to use cross-image features which describe
the relation between two faces, thus better fits the face
verification problem. Note that our observation can be generalized
to verification of any type of objects. For instance, given two
images of animals, we can verify whether both images are for cats
by running each of them separately through a cat detector trained
on images of cats vs non-cats, then combine the detector's
confidences. This is inherently different from using a
verification classifier which receives a pair of images and
classifies them jointly as being both cats or not. The latter
classifier would be trained on cross-image features extracted from
cat-cat pairs vs cat-noncat pairs, which is more discriminative
for object similarity and dissimilarity.

Our cross-image features are embarrassingly simple, we use a large
bank of normalized correlation filters between patches across the
pair of images at different sizes, locations, and orientations.
Additionally, we use Haar-like features similar to the ones used
in \cite{VIOLA}, however, obtained across the pair of images.
Consequently, we use AdaBoost to weight and select the most
discriminative filters. Our method is derived from Viola and Jones
face detector \cite{VIOLA}; however, instead of using filters
within an individual image, we use across-image filters such that
we capture the relation between the pair of images, rather than
capturing individual image features. It is important to note that
cross-image features are extracted from two images in order to
encode their similarity. This is inherently different from
relating pairs or triplets of features extracted from the same
image, which is typically used to encode the spatial relations
between the features as in for example \cite{Gu2010} and
\cite{Ni2009}. The rest of the paper is organized as follows: In
the next section, we present our proposed cross-image features,
followed by AdaBoost classifier training. The experiments and the
results are described in Section $3$. Finally, Section $4$
concludes the paper.

\section{Proposed Method}
\begin{figure}
   \includegraphics[width=1\linewidth]{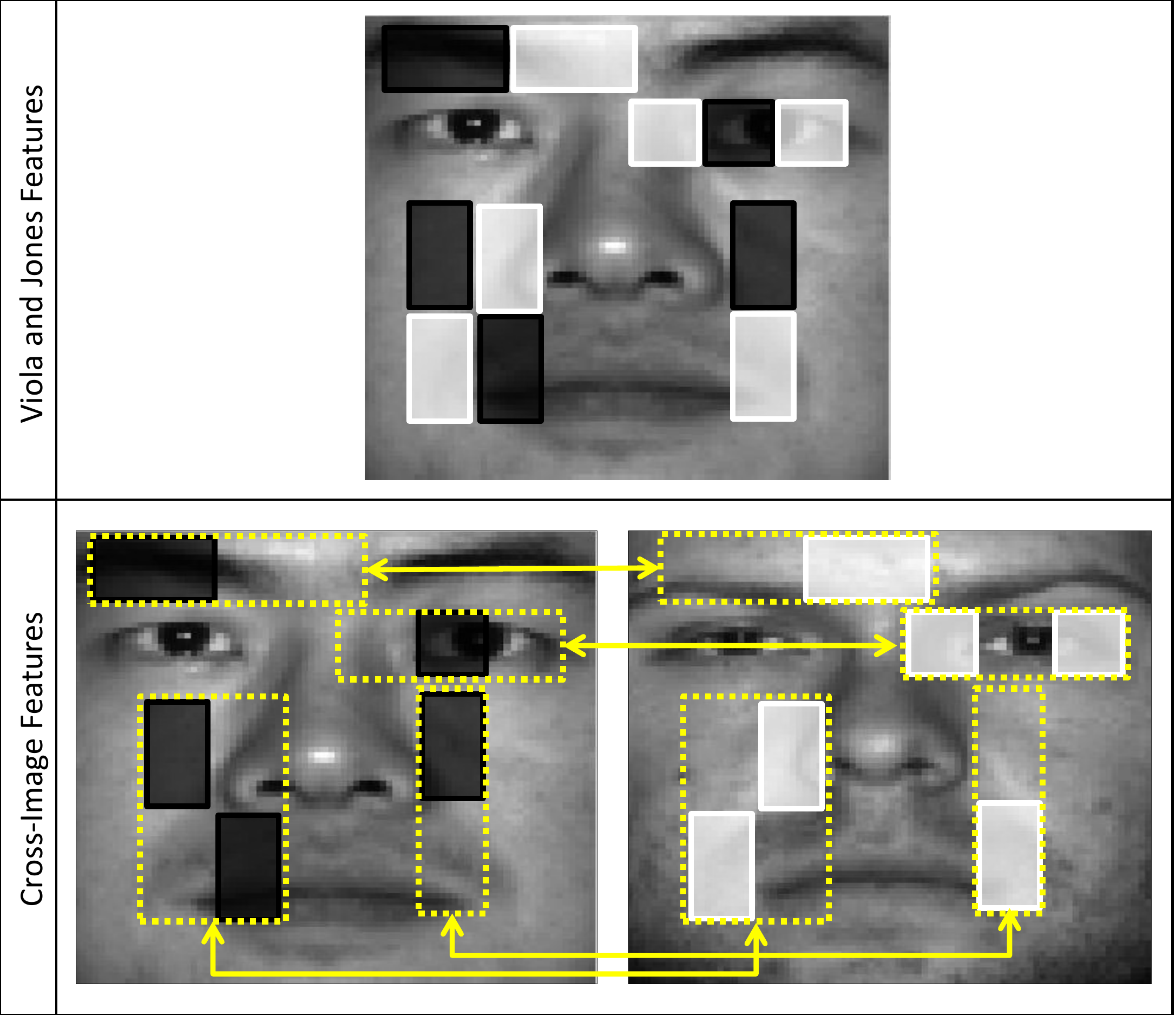}
   \caption{The top row shows the four rectangle features from
   \cite{VIOLA}. The bottom row shows the corresponding cross-image features, which are similar to the
   original Haar-like features except that the white part of the filters are obtained from the
   second image in order to capture the difference between the two images in the pair.
} \label{fig:feat}
\end{figure}

Our cross-image features for face verification are based on the
simple rectangle filters presented by Viola and Jones
\cite{VIOLA}. However, we extend the features to operate across
pairs of images rather than within individual images. While these
features seem simple, the experiments demonstrate their superior
discriminative capabilities in face verification. Figure
\ref{fig:feat} illustrates the difference between the features of
\cite{VIOLA} and ours. We capture the differences between the two
images, instead of the variation within the image. In particular,
given a pair of face images $I_{1}$ and $I_{2}$, which we aim to
classify as belonging to the same identity or not, let $r_{i}$
denote a box $i$ positioned at location $(x_{i},y_{i})$, with
width and height $(w_{i},h_{i})$, and orientation $\theta_{i}$. We
define two types of filters:

\begin{itemize}

\item{Haar-like cross-image filters:} This type of filter compares
the box rectangular sum between the image pair

\begin{equation}\label{eq1}
f_{i} = \sum_{(x,y) \in r_{i}}I_{1}(x,y) - \sum_{(x,y) \in
r_{i}}I_{2}(x,y).
\end{equation}

We use four versions of this type similar to \cite{VIOLA};
however, the features are obtained across the image pair as shown
in figure \ref{fig:feat}. The black part of the rectangle boxes is
obtained from the first image, while the white part is obtained
from the second.

\item{NCC cross-image filters:} This type of filter computes the
normalized cross correlation (NCC) of the rectangular box between
the image pair

\begin{equation}\label{eq2}
f_{j} = \sum_{(x,y) \in r_{i}} \frac{(I_{1}(x,y) -
\bar{I_{1}})(I_{2}(x,y) - \bar{I_{2}}) }{\sigma_{1}\sigma_{2}}
\end{equation}

\end{itemize}

We use a single version of this feature where the NCC is obtained
between corresponding patches in the pair at the same location
(i.e. both black and white boxes are placed at the same spatial
coordinates in the pair). Note that since the correlation is
normalized, these features are robust against illumination
changes.

We quantize all possible positions, sizes, and orientations for
each of the filters and obtain about $25,000$ features for each
pair of images. Calculating the cross-image features for thousands
of image boxes is time consuming. Therefore, in order to rapidly
compute the features, we adopt the integral image method
\cite{VIOLA} and apply it on the cross-image features. The
integral image $\delta$ for an image $I$ is defined as

\begin{equation}\label{eq3}
\delta(x,y) = \sum_{x' \leq x , y' \leq y}I(x',y').
\end{equation}

Each box summation in $I$ can be obtained using four anchors from
$\delta(x,y)$ similar to \cite{VIOLA}. For the Haar-like
cross-image filters, we first obtain the integral image for each
of the images in the pair. Consequently, every box summation is
obtained from its corresponding integral image. On the other hand,
in order to use the integral image for efficient computation of
the NCC cross-image features, we first expand equation \ref{eq2}
and apply a few simple operations to reformulate it as

\begin{equation}
    \label{eq4}
    f_{j}=\frac{n\sum{I_{1}I_{2}}-\sum{I_{1}}\sum{I_{2}}}{\sqrt{(n\sum{I_{1}^2}-\sum^2{I_{1}})(n\sum{I_{2}^2}-\sum^2{I_{2}})}},
\end{equation}where the summation is over
all the pixels in the box filter, and $n$ is the number of pixels.
Consequently, we obtain five integral images corresponding to each
of the terms, in particular, $(I_{1}, I_{2}, I_{1}I_{2},
I_{1}^{2},I_{2}^{2})$. Using these internal images, each of the
summation terms in equation \ref{eq4} is efficiently computed
using four anchors from the corresponding integral image.

In the training process, we use AdaBoost to select a subset of
features and construct the classifier. In each round, the learning
algorithm chooses from a heterogenous set of filters, including
the Haar-like filters and the NCC filters. The AdaBoost algorithm
also picks the optimal threshold for each feature. The output of
AdaBoost is a classifier which consists of a linear combination of
the selected features. For details on AdaBoost, the reader is
referred to \cite{VIOLA}.

\section{Experiments}

We extensively experimented on the proposed ideas using three
standard datasets: Extended Yale B \cite{Georghiades2001}, CMU PIE
\cite{CMUPIE}, and Labeled Face in the Wild (LFW)
\cite{Huang2007}. Figure \ref{fig:selected} shows the the highest
weighted rectangle features obtained after boosting of the
cross-image features for theses three datasets. Figure
\ref{fig:result} shows example verification results from the three
datasets. In all experiments, we report our performance using the
accuracy at EER ($100\%$-ERR) similar to \cite{Schroff2011}, where
EER is the equal error rate, which is the average value of the
false accept rate (FAR) and the false reject rate (FRR).

\begin{itemize}

\item {\textbf{Extended Yale B}}: This is a standard face database
consisting of $2414$ frontal-face images of $38$ individuals. The
face images are normalized to the size of $192 \times 168$. These
face images are captured under different laboratory-controlled
lighting conditions \cite{Lee2005}. There are about $64$ images
for each individual. We follow a standard experimental setup
similar to \cite{Wright2009} and randomly select half of the
images of each subject for training and the other half for
testing. Figure \ref{fig:accuracy} top left shows the obtained ROC
curve for Extended Yale B dataset. We compared our method with
state-of-the-art methods in table \ref{table:methods_comparison1}
and figure \ref{fig:accuracy}. It is clear that our cross-image
features outperforms the other approaches.

\end{itemize}

\begin{table}[htb]
\caption{Face verification results for Extended Yale B and CMU PIE
datasets.} \label{table:methods_comparison1}
\begin{tabular}{lccc}
\\
 \hline
 Method & \multicolumn{3}{l}{Accuracy at EER (\%)} \\
 \cline{2-4}
  & YaleB & PIE Illum& PIE Light \\
 \hline
 Heusch et al. \cite{Heusch2006} & 73.64 & 84.85 & 89.63 \\
 Zhang et al. \cite{Zhang2009} & 85.09 & 79.40 & 82.77 \\
 WLBP-HS \cite{Goh2011} & 88.46 & 86.80 & 90.07 \\
 WLFuse \cite{Goh2011}  &  91.25 & 86.89 & 90.83 \\
 Our method & \textbf{95.70} & \textbf{92.49} & \textbf{98.61}\\
 \hline
 \end{tabular}
\end{table}
\begin{itemize}

\begin{figure}
   \includegraphics[width=1\linewidth]{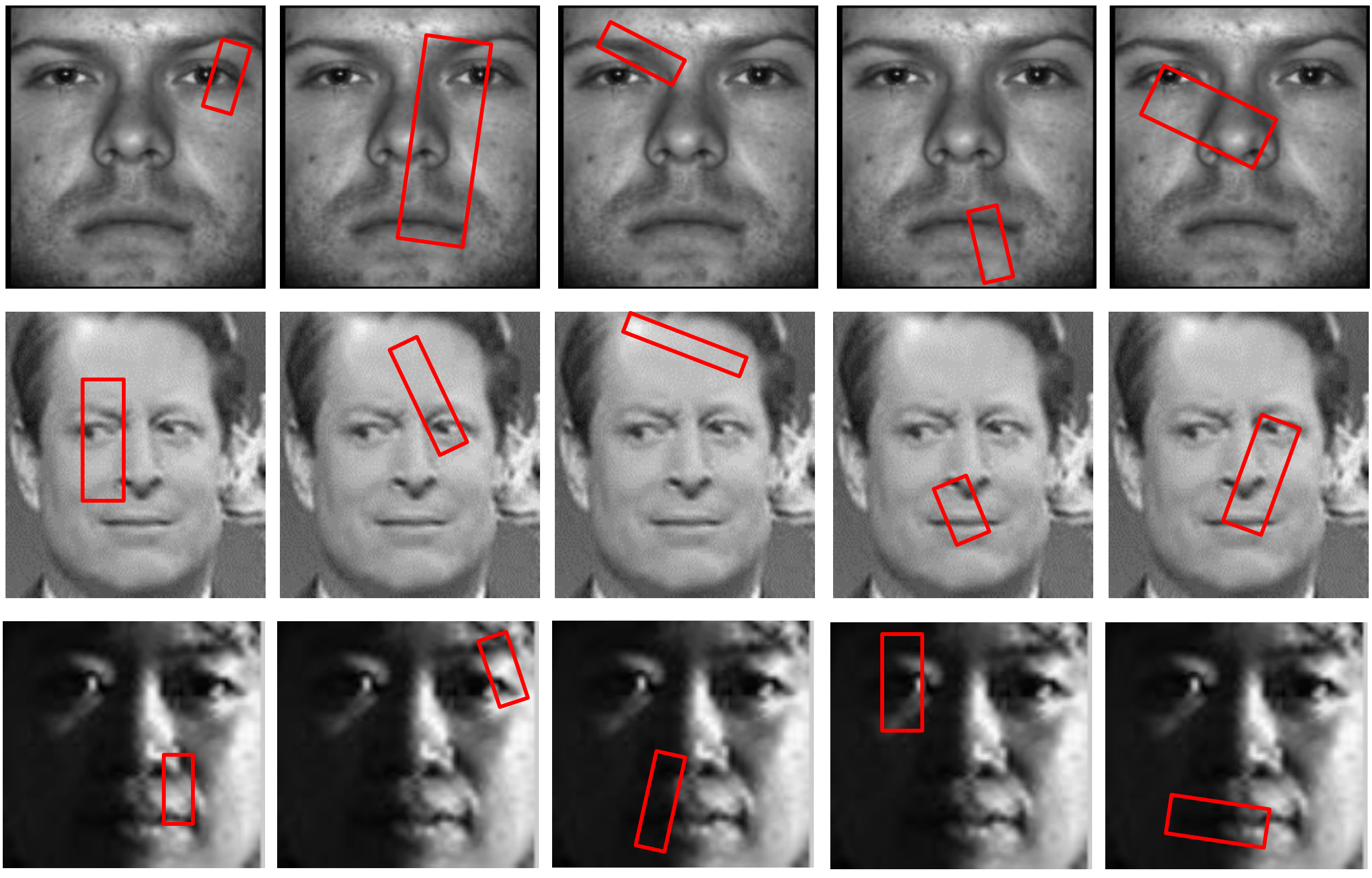}
   \caption{The highest weighted rectangle features obtained after Boosting for Extended Yale B dataset (top), LFW dataset (middle), and CMU PIE dataset(bottom).
} \label{fig:selected}
\end{figure}

\begin{figure}
   \includegraphics[width=1\linewidth]{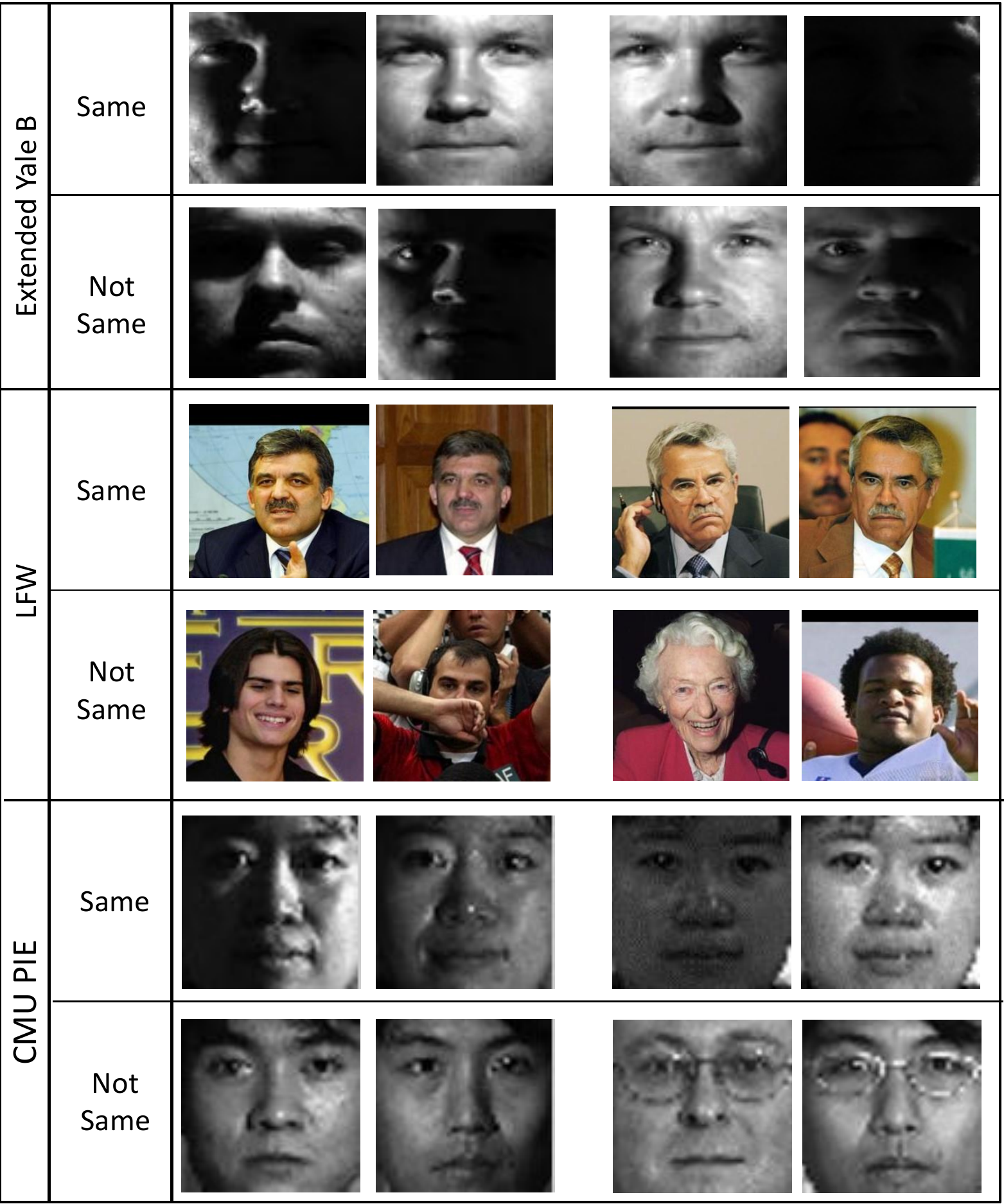}
   \caption{Example face verification results from Extended Yale B, LFW, and CMU PIE datasets.
} \label{fig:result}
\end{figure}

\begin{figure}
   \includegraphics[width=1\linewidth]{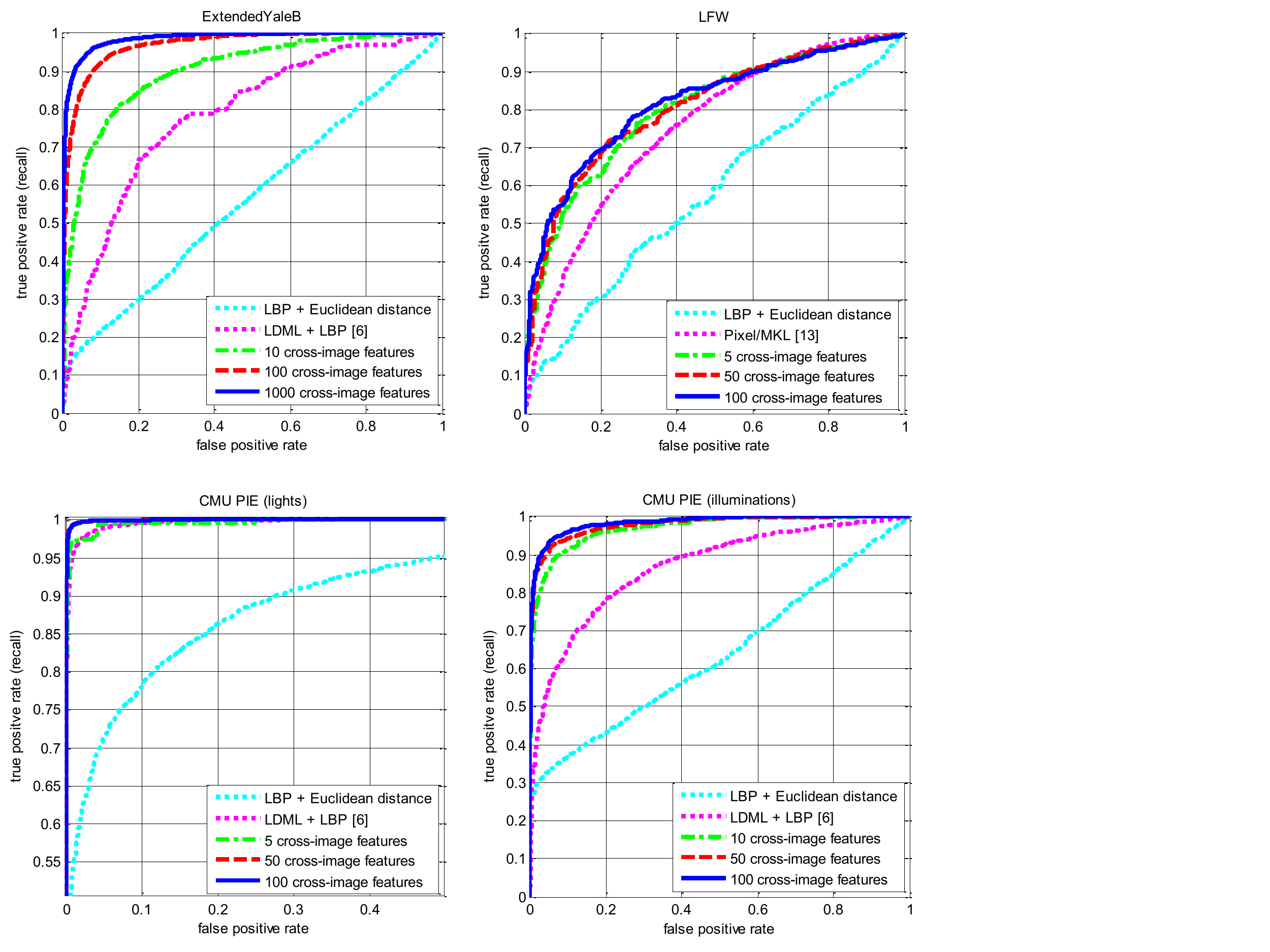}
   \caption{The performance of our method using the highest weighted cross-image features. In the ROC curves, we compare our method with
   LBP + Euclidean distance, where the face similarity is measured by the distance between LBP features extracted from the faces.
   Additionally, we compare our performance with the methods from [6] and \cite{Pinto2009}.
} \label{fig:accuracy}
\end{figure}

\item {\textbf{CMU PIE}}: This dataset contains $68$ subjects with
$13$ different poses and $4$ different expressions under $43$
different lighting conditions. We follow the experiment setup in
\cite{Goh2011} and use two subsets from this database, namely,
``illumination'' (without ambient light) and ``lighting'' (with
ambient light). Similar to \cite{Goh2011}, in the illumination
subset we use $3$ images per person for training, and another $8$
images with different illumination conditions are randomly
selected for testing. In the lighting subset, we use 5 images for
training and 10 for testing. Table \ref{table:methods_comparison1}
shows the performance of our method compared to several other
approaches. Additionally, figure \ref{fig:accuracy} bottom left
and right shows the obtained ROC curve for CMU PIE lighting (left)
and illumination (right).

\item {\textbf{Labelled Faces in the Wild}}: LFW dataset contains
$13233$ face images collected from the web. The database includes
$5749$ individuals, $1680$ of them have two or more distinct
photos. This dataset is very challenging due to the variation in
illumination and pose; therefore, it is useful for comparing the
effectiveness of different low-level feature descriptors
\cite{Schroff2011}. In our experiments, we use the aligned version
of LFW, and follow a standard $10$-fold cross validation suggested
by the authors of \cite{Huang2007}. Figure \ref{fig:accuracy} top
right shows the obtained ROC curve for LFW dataset. Moreover,
table \ref{tbl:features_comparison2} compares our cross-image
features with the state-of-the-art feature descriptors. It is
clear that our method is robust to the challenging factors in LFW,
and therefore outperforms other features.

\begin{table}[htb]
\caption{Comparison of the accuracy at EER for different feature
descriptors on LFW dataset} \label{tbl:features_comparison2}
\begin{tabular}{r||c|c|c|c}
 \hline
 Feature & TPLBP \cite{Goh2011}& SIFT & look-alike \cite{Schroff2011} & ours \\
 \hline
 Accuracy(\%) & 69.2 & 69.1 & 70.8 & \textbf{75.4} \\
 \hline
 \end{tabular}
\end{table}

\end{itemize}

\section{Conclusion}

We proposed a new robust features for face verification based on
cross-image similarity. Our approach extracts simple rectangle
features from the pair of images jointly, thus capturing
discriminative properties of the pair. Through experiments, we
demonstrated the power of our proposed approach on challenging
datasets. In the future, we will explore the extension of our
cross-image features for general object verification problem.


%
\bibliographystyle{abbrv}
\bibliography{sigproc}  
%
%

\end{document}